\documentclass[letterpaper, 10 pt, conference]{ieeeconf}

\IEEEoverridecommandlockouts

\usepackage{amsmath} 
\usepackage{amssymb}  
\usepackage{cite}
\usepackage{bm}
\usepackage{graphicx}
\usepackage{xcolor}
\usepackage{booktabs}
\usepackage{multirow}
\usepackage{makecell}
\usepackage{subcaption}
\usepackage{diagbox}

\title{\LARGE \bf
PCHC: Enabling Preference Conditioned Humanoid Control via Multi-Objective Reinforcement Learning
}

\author{
    Huanyu Li$^{1}$, 
    Dewei Wang$^{2}$, 
    Xinmiao Wang$^{3}$, 
    Xinzhe Liu$^{4}$, 
    Peng Liu$^{1\dagger}$, 
    Chenjia Bai$^{5\dagger}$, 
    Xuelong Li$^{5}$%
    \thanks{$^{1}$Harbin Institute of Technology}%
    \thanks{$^{2}$University of Science and Technology of China}%
    \thanks{$^{3}$Harbin Engineering University}%
    \thanks{$^{4}$ShanghaiTech University}%
    \thanks{$^{5}$Institute of Artificial Intelligence (TeleAI), China Telecom}%
    \thanks{$^{\dagger}$Corresponding authors}%
}

\begin{document}
\maketitle
\thispagestyle{empty}
\pagestyle{empty}

\begin{abstract}

Humanoid robots often need to balance competing objectives, such as maximizing speed while minimizing energy consumption. While current reinforcement learning (RL) methods can master complex skills like fall recovery and perceptive locomotion, they are constrained by fixed weighting strategies that produce a single suboptimal policy, rather than providing a diverse set of solutions for sophisticated multi-objective control.
In this paper, we propose a novel framework leveraging Multi-Objective Reinforcement Learning (MORL) to achieve Preference-Conditioned Humanoid Control (PCHC). Unlike conventional methods that require training a series of policies to approximate the Pareto front, our framework enables a single, preference-conditioned policy to exhibit a wide spectrum of diverse behaviors. To effectively integrate these requirements, we introduce a Beta distribution-based alignment mechanism based on preference vectors modulating a Mixture-of-Experts (MoE) module.
We validated our approach on two representative humanoid tasks. Extensive simulations and real-world experiments demonstrate that the proposed framework allows the robot to adaptively shift its objective priorities in real-time based on the input preference condition.
\end{abstract}

\section{Introduction}

Humanoid robots represent a premier embodied platform for replacing humans in hazardous environments and labor-intensive tasks. 
While Model Predictive Control (MPC) has traditionally enabled locomotion across diverse terrains \cite{gao2024mpc, kim2007humanoidmpc}, Reinforcement Learning (RL) \cite{kaelbling1996reinforcement}, empowered by high-performance parallel simulators \cite{zakka2025mujoco, makoviychuk2021isaac}, has recently achieved unprecedented robustness in complex control scenarios. 
Current RL frameworks, particularly those utilizing policy gradient methods \cite{PPO1, PPO2}, have successfully addressed challenging tasks including fall recovery \cite{standup, huang2025host}, blind locomotion over uneven ground \cite{gu2024advancing, cui2024adapting, xue2025unified}, vision-based locomotion \cite{wang2025more, long2025learning, wang2025beamdojo, zhuang2024humanoid}.
Despite these advancements, these methods typically rely on predefined weights to scalarize competing reward objectives, ultimately collapsing the multi-faceted nature of the task into a rigid, single-objective policy.

\begin{figure}[t]
\centering
\includegraphics[width=1.0\linewidth]{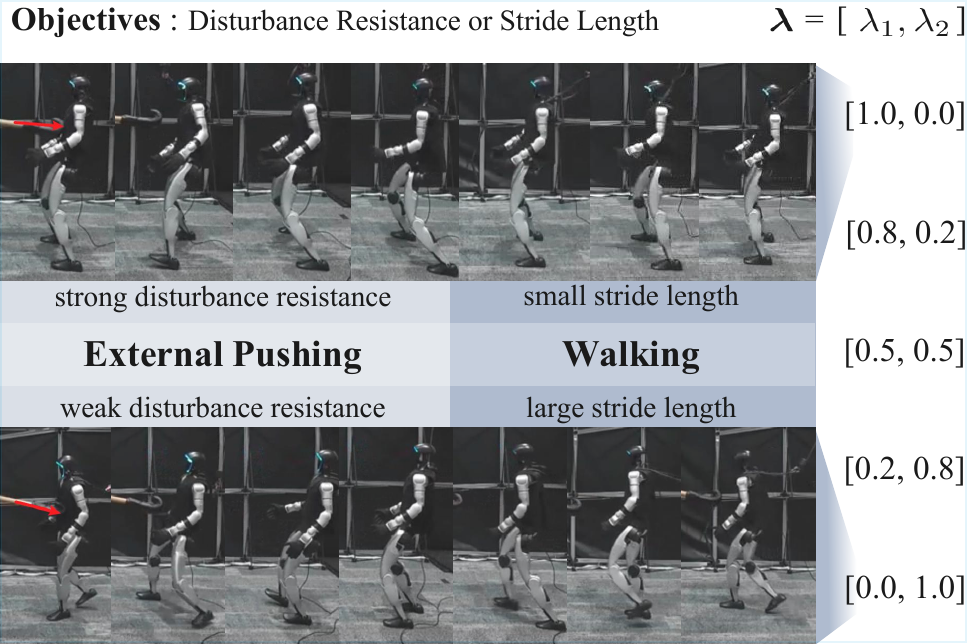}
\caption{\textit{PCHC} integrates MORL with humanoid control tasks which enables the robot perform different multi-objective preference-aligned behaviors given preference vector $\bm \lambda$. For example, \textit{PCHC} enables the robot to balance stride length and disturbance resistance in the humanoid walking task.}
\label{fig:walking_performance}
\vspace{-1.5em}
\end{figure}

In the real world, many complex control problems, such as humanoid robot locomotion \cite{gu2025humanoid}, intrinsically involve multiple conflicting objectives that cannot be optimized simultaneously. Single-objective RL approaches typically rely on manually tuned, fixed reward weights to balance these competing objectives, yielding static behaviors that cannot dynamically adapt to changing environmental demands or shifting user priorities during deployment. 
Multi-objective reinforcement learning (MORL) \cite{van2014multi, hayes2021practical}, which aims to approximate the Pareto front \cite{navon2020learning}, provides a highly suitable solution. While many existing Q-learning \cite{reymond2022pareto, basaklar2022pd, yang2019generalized} and policy gradient \cite{cai2023distributional, terekhov2024search, xu2020prediction} methods have been developed, extending these paradigms to high-dimensional continuous humanoid control remains a formidable challenge, as it requires balancing the explosion of the state-action space with the intricate, multi-objective trade-offs inherent in complex dynamics.
This dynamic adaptability is especially critical for humanoid robots, which must reconcile competing priorities ranging from speed and energy efficiency in industrial logistics to stability and agility in search-and-rescue operations. Despite its potential, the application of MORL to humanoid locomotion remains non-trivial. Unlike standard MORL benchmarks that rely on simplified environments \cite{alegre2022mo}, humanoid control involves high-dimensional dynamics and a multitude of dense, manually crafted reward functions \cite{huang2025host, wang2025more, gu2024advancing}. Consequently, developing a unified multi-objective humanoid control policy capable of navigating these complex Pareto trade-offs in real-time remains an unresolved challenge.

In this paper, we propose \textit{PCHC}, a novel framework that enables a single, unified policy to achieve versatile humanoid control while satisfying multiple preference-conditioned objectives. We formulate preference-conditioned humanoid control as a Multi-Objective Reinforcement Learning (MORL) problem, aiming to approximate the Pareto-optimal front. To eliminate the overhead of policy switching, PCHC trains a preference-conditioned policy that employs linear scalarization across diverse objectives.
Given that RL-based humanoid control typically involves an extensive array of reward functions, we distinguish between objective-specific rewards and auxiliary rewards. Specifically, only the rewards directly relevant to target-specific objectives are subjected to preference-based scalarization during training, while auxiliary rewards remain invariant to ensure fundamental stability. To effectively map preference vectors to robot behaviors, we introduce a multi-expert network architecture where expert routing is governed not by a gating network, but by a Beta distribution parameterized by the preference vector. Each expert implicitly acquires preference-aligned knowledge, and the entire ensemble is trained to approximate the Pareto front through a guided, weighted aggregation.

We evaluate \textit{PCHC} on two humanoid tasks: walking and fall recovery, both characterized by intrinsically conflicting objectives. Specifically, we investigate the trade-offs between disturbance resistance versus stride length in walking, and disturbance resistance versus energy efficiency in fall recovery. Experimental results demonstrate that \textit{PCHC} can dynamically shift its behavioral priorities across the Pareto front based on the input preference vector. Our primary contributions are summarized as follows:
\begin{itemize}{}{}
    \item We propose \textit{PCHC} that integrates MORL with humanoid control to obtain distinct policy outputs conditioned on different preferences over objectives.
    \item We design a multi-expert module combined with a preference-parameterized Beta distribution to perform preference injection. 
    \item We validate the proposed framework on two humanoid robot tasks and demonstrate its effectiveness across multiple metrics through extensive simulation and real-world experiments.
\end{itemize}

\section{Related Work}
\subsection{Humanoid Control with Reinforcement Learning}
RL algorithms \cite{kaelbling1996reinforcement, PPO1} are now widely applied to humanoid robot control, greatly unlocking the potential of humanoid robots \cite{cui2024adapting, xue2025unified,wang2025more, long2025learning, shi2025adversarial, gu2024advancing, zhuang2024humanoid, wang2025beamdojo, fu2024humanplus, ben2025homie}. 
Humanoid robots are capable of walking on uneven terrains \cite{gu2024advancing, cui2024adapting}, while the use of external sensors such as LiDAR or depth cameras enables them to handle more complex scenarios with improved stability \cite{wang2025more, long2025learning, wang2025beamdojo}. By training policies to track human motion, humanoid robots can perform tasks such as dancing through real-time tracking of human body poses \cite{he2025asap, xie2025kungfubot, fu2024humanplus}. Fall recovery \cite{huang2025host, standup} and walking \cite{gu2024humanoid, shi2025adversarial} are fundamental and essential skills for humanoid robots, Host \cite{huang2025host} employs multi-critic RL \cite{mysore2022multi} and a multi-stage framework to achieve recovery to standing from various falling postures and terrains. Walking on complex terrains has been addressed through reward function design, policy distillation, and privileged information reconstruction \cite{gu2024advancing, wang2025more, zhuang2024humanoid, long2025learning}.
Most of these approaches rely on training in simulation \cite{makoviychuk2021isaac, zakka2025mujoco} and employ techniques such as domain randomization to mitigate the sim-to-real gap. However, existing works fail to accommodate multiple, and often conflicting, objectives under varying preference conditions which would enable policies to achieve better adaptability across different situations.

\subsection{Multi-Objective Reinforcement Learning}
Existing MORL algorithms can be classified as multi-policy and single policy. Multi-policy approaches \cite{reymond2022pareto, yang2019generalized} seek to learn a collection of policies that collectively approximate the Pareto front. In contrast, single-policy methods \cite{van_moffaert_scalarized_2013, basaklar2022pd, terekhov2024search} convert a multi-objective problem into a single-objective one by aggregating multiple rewards into a scalar value using a scalarization function. In addition, early works \cite{reymond2022pareto, basaklar2022pd, yang2019generalized} primarily follow the Q-learning paradigm, where the agent learns to approximate the value function for different preference weights. Recently, with the success of policy-gradient algorithms in continuous control, some studies have extended them to MORL by incorporating objective weights \cite{terekhov2024search, xu2020prediction} or distributional value function into the policy optimization process \cite{cai2023distributional}. 
Meanwhile, MORL has gradually been adopted in fine-tuning of large language models (LLMs) \cite{rame2023rewarded, jang2023personalized} to balance multiple reward signals. This also offering insights into preference-conditioned optimization strategies applicable to robotics.
Existing works on MORL for robot control have remained in toy scenarios like MO-Gym \cite{alegre2022mo}. In this work, we extend MORL to complex humanoid control tasks and validate our approach through both simulation and real-world experiments.

\section{Method}

\subsection{Preliminaries}
We formulate the humanoid control problem as an RL problem, which can be modeled as a Markov Decision Process (MDP) formally described by a tuple \((S, A, P, R, \gamma, \rho_0)\). Here, 
\(S\) denotes the state space, 
\(A\) represents the action space, 
\(P\) is the state transition probability, 
\(R\) is the reward function, 
\(\gamma\) is the discount factor, and 
\(\rho_0\) is the initial state distribution.
While standard RL focuses on a single reward signal, many real-world tasks involve multiple, often conflicting objectives. This motivates MORL, which extends the standard MDP to a multi-objective MDP (MOMDP). 

\begin{figure}[t]
\centering
\includegraphics[width=1.0\linewidth]{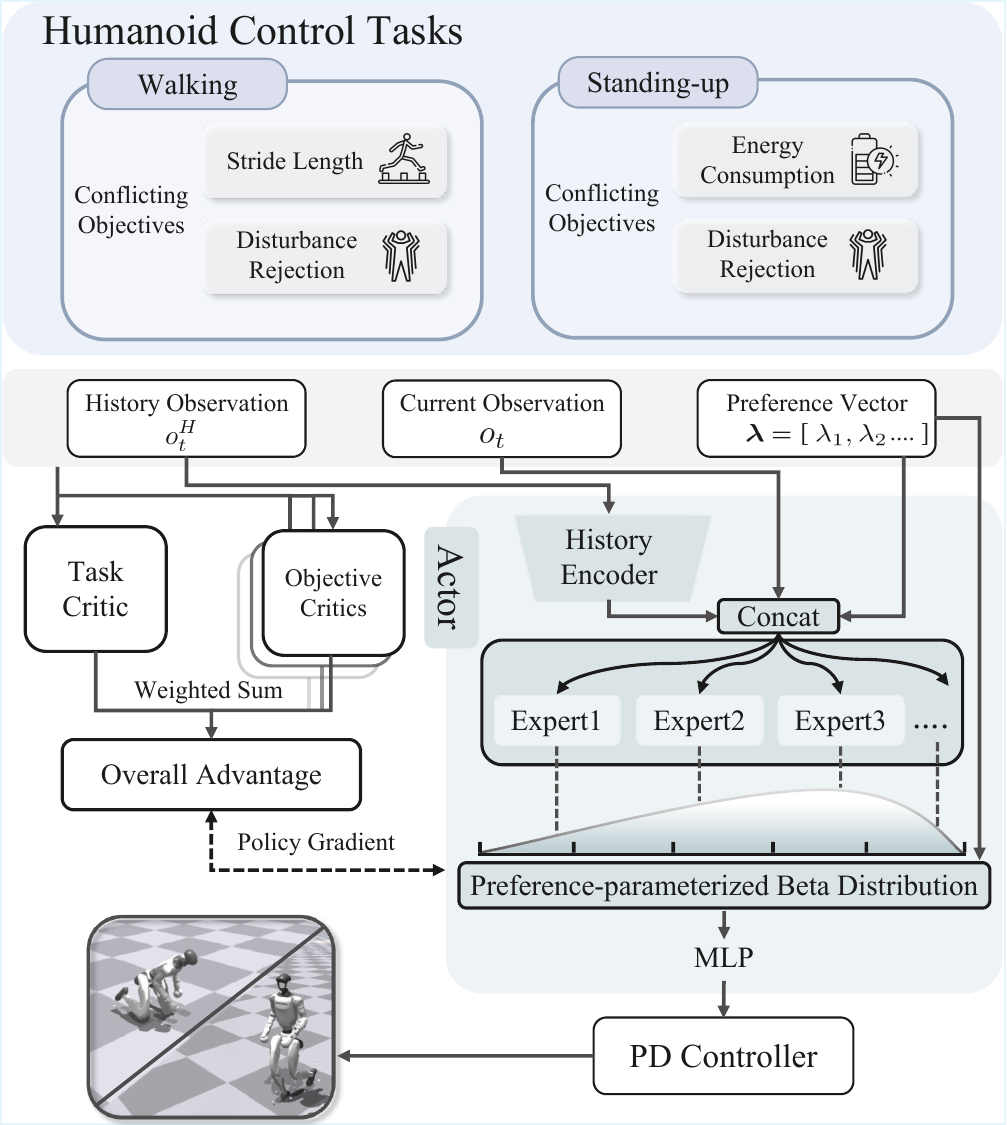}
\caption{Overview of the proposed \textit{PCHC} framework. \textit{PCHC} adopts multiple critics and employs a \textit{Preference Condition Injection} module based on a preference-parameterized Beta distribution achieving multi-objective control on two humanoid control tasks. }
\label{fig:method}
\end{figure}

In MOMDPs, the reward function 
$R: S \times A \rightarrow \mathbb{R}^m$
returns a reward vector 
$\mathbf{r} = (r_1, r_2, \dots, r_m)$, 
where \(m\) denotes the number of objectives. The value function of a policy $\pi$ is defined as
\begin{equation}
\mathbf{v}^{\pi}(s) \triangleq \mathbb{E}_{\pi}\left[\sum_{i=0}^{\infty} \gamma^{i} \mathbf{r}_{t+i} \mid S_{t}=s\right].
\end{equation}
To compare two policies in a multi-objective setting, their values are evaluated across all objectives. 
A policy $\pi_1$ is said to dominate another policy $\pi_2$ if and only if it improves at least one objective without reducing the values of the others. As a result, the set of undominated policies forms the Pareto front:
\begin{equation}
PF(\Pi)=\{\pi\in\Pi\mid\nexists\pi^{\prime}\in\Pi:\mathbf{v}^{\pi^{\prime}}\succ_P\mathbf{v}^\pi\},
\end{equation}
where $\succ_P$ is the Pareto dominance relation:
\begin{equation}
\mathrm{v}^\pi\succ_P\mathrm{v}^{\pi^{\prime}}\Longleftrightarrow(\forall i:\mathrm{v}_i^\pi\geq\mathrm{v}_i^{\pi^{\prime}})\wedge(\exists i:\mathrm{v}_i^\pi>\mathrm{v}_i^{\pi^{\prime}}),
\end{equation}
and $\Pi$ is the set of all possible policies.

For complex humanoid control tasks, obtaining the exact Pareto front is practically infeasible. Therefore, the goal of multi-objective optimization is to approximate the Pareto front as closely as possible. To evaluate the approximated Pareto front, two commonly used metrics are considered: hypervolume, which reflects the coverage of the objective space, and sparsity, which indicates the diversity of solutions. Let \(P = \{p_1, p_2, \dots, p_n\} \subset \mathbb{R}^m\) denote an approximation of Pareto front containing \(n\) solutions, where each solution \(p_i\) corresponds to the values of \(m\) objectives. The hypervolume indicator quantifies the volume of the objective space dominated by the Pareto front. Formally, it is defined as
\begin{equation}
H(P) = \Lambda \Big( \bigcup_{p \in P} [p, r] \Big),
\end{equation}
where \(r \in \mathbb{R}^m\) is a reference point, \([p, r]\) denotes the hyperrectangle spanning from \(p\) to \(r\) across all dimensions, and \(\Lambda(\cdot)\) represents the Lebesgue measure (i.e., the volume).
For the same P, sparsity is defined as:
\begin{equation}
S(P) := \frac{1}{n-1} \sum_{j=1}^{m} \sum_{i=1}^{n-1} \left(P_{ij} - P_{i+1\,j}\right)^2
\end{equation}
where \(P_i\) is the \(i^{\text{th}}\) solution in \(P\) and \(P_{ij}\) means the value \(j^{\text{th}}\) objective. It is evident that a larger hypervolume indicates a better approximation of the Pareto front, while lower sparsity values reflect a denser distribution of solutions.

\begin{figure}[t]
\centering
\includegraphics[width=1.0\linewidth]{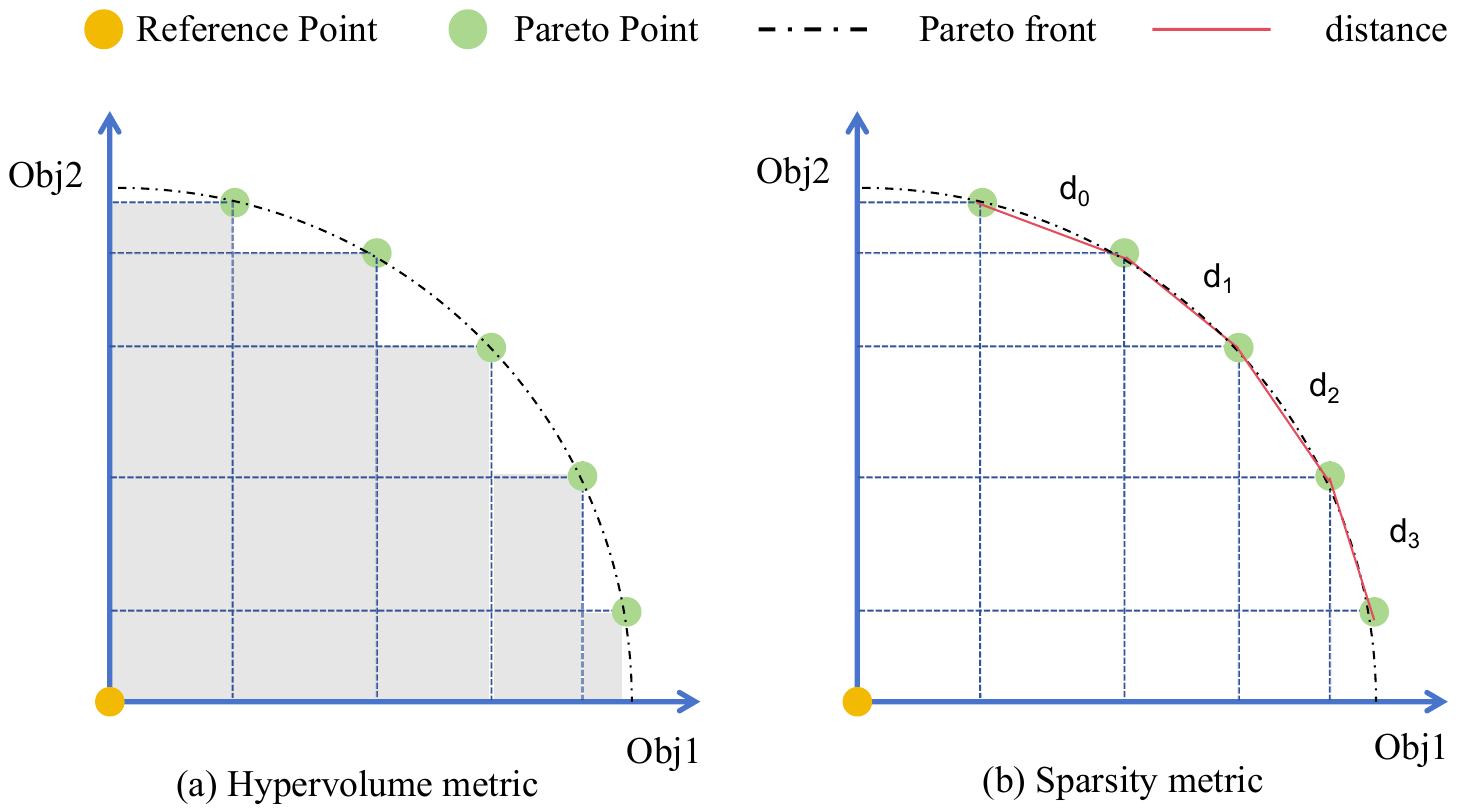}
\caption{In two-objective space: (a) Hypervolume is represented with the shaded area bounded by the Pareto points and the reference point. (b) Sparsity measures the average square distance between consecutive points.}
\label{fig:metrics}
\end{figure}

\subsection{Preference-Conditioned Humanoid Control}
An overview of the proposed framework and conflicting objectives of two humanoid control tasks are shown in Fig.~\ref{fig:method}.
Unlike multi-policy approaches that train a set of policies to approximate the Pareto front, we adopt a single-policy formulation in order to dynamically adapt the policy. To achieve this, we condition the policy on a preference vector that specifies the relative importance of multiple objectives. 
Formally, we define a preference vector $ \bm \lambda \in \mathbb{R}^m$, where each entry $\lambda_i \geq 0$ corresponds to the weight of objective $i$, and the vector lies on the probability simplex, i.e., $\sum_{i=1}^m \lambda_i = 1$. 
During training, $\bm \lambda$ is sampled from a Dirichlet distribution and concatenated with the other observation of the robot.

A key challenge in preference-conditioned training is accurate value estimation under dynamic preference vector. Inspired by previous methods \cite{huang2025host, mysore2022multi}, we adopt multiple critics to estimate returns, where one critic is assigned to evaluate the task-related return and additional critics are assigned to estimate each objective-specific return.
We employ GAE \cite{schulman2015high} to estimate the advantage for each critic, and then compute a weighted sum to obtain the overall advantage.
The final preference-conditioned advantage is thus obtained as
\begin{equation}
\hat{A}^{\bm \lambda} = \Big( \sum_{i=1}^m \lambda_i \, \hat{A}_i^{\rm obj} \Big) \;+\; \hat{A}^{\text{task}},
\end{equation}
where $\hat{A}_i^{\rm obj}$ and $\hat{A}^{\text{task}}$ denote the normalized advantage for objective-related and task-related rewards, respectively. This formulation enables the policy update to dynamically reflect the user-specified trade-off among conflicting objectives while consistently enforcing essential task constraints.
The preference-conditioned policy is optimized using Proximal Policy Optimization (PPO) \cite{PPO1}. 
Specifically, with a mimibatch $\{(s_k, a_k, \hat{A}^\lambda_k)\}_{k}$, the loss of the actor is defined as: 
\begin{equation}
\mathcal{L}_\text{}(\theta) = 
\sum_{k} 
\min \Big( 
r_k(\theta) \, \hat{A}^{\bm\lambda}_k,
\text{clip}(r_k(\theta), 1 - \epsilon, 1 + \epsilon) \, \hat{A}^{\bm \lambda}_k
\Big),
\end{equation}
where $r_k(\theta) = \frac{\pi_\theta(a_k \mid s_k, \bm \lambda)}{\pi_{\text{ref}}(a_k \mid s_k, \bm \lambda)}$, $\pi_{\text{ref}}$ is the policy directly after sampling the trajectory.

\subsection{Preference Condition Injection}
Directly concatenating the preference vector $\bm \lambda$ with other observation as policy input is often insufficient, since the mapping from preference condition to low-level motor behaviors in humanoid control is highly non-linear and complex. 
To address this, we propose a \textit{Preference Condition Injection} module, which consists of multiple experts and a Beta distribution parameterized by the preference vector.
Instead of relying on a gating network to determine expert routing, we parameterize a Beta distribution using the preference vector, and sample routing weights from this distribution. For two conflicting objectives, the preference vector $\lambda$ lies on a 1D probability simplex. The Beta distribution is uniquely suited for this architecture as the equivalent of a two-dimensional Dirichlet distribution. Its domain naturally aligns with the $[0,1]$ interval of interpolating between two conflicting objectives, providing mathematically bounded and smooth density curves that enforce a structured interpolation between experts along the Pareto front.

Given a preference vector $\bm \lambda = [\lambda_1, \lambda_2]$, 
we define the Beta distribution parameters as
$\alpha = \alpha_0 + \lambda_1 \cdot s, \beta = \beta_0 + \lambda_2 \cdot s,$ where $\alpha_0,\beta_0 > 0$ are base offsets and $s$ is a scaling factor.  
For each expert branch $j \in \{1,\dots,N_{B}\}$ with fixed position $x_j \in [0,1]$, 
the unnormalized weight is computed by evaluating the Beta density:

\begin{equation}
\tilde{w}_j = \frac{x_j^{\alpha-1} (1-x_j)^{\beta-1}}{B(\alpha,\beta)},
\end{equation}
where $B(\alpha, \beta) = \int_0^1 t^{\alpha-1} (1-t)^{\beta-1} \, dt$ is the Beta function. The final routing weights are first normalized.
Each expert network receives the current observation as well as the output from the history encoder as shown in Fig. \ref{fig:method}. Their outputs are aggregated through a weighted sum according to the routing weights $\tilde{w}_j$.
Each expert in the multi-expert actor network implicitly learns a distinct solution on the Pareto front. By determining their routing weights through a preference-parameterized Beta distribution, the framework can leverage the interpolation of these expert network outputs to better generate solutions at other points on the Pareto front. To ensure a comprehensive representation of the Pareto front without introducing excessive complexity, we set the number of experts to $N_{B}=5$ in our implementation. Their fixed positions $x_j$ are evenly distributed at 0.0, 0.25, 0.5, 0.75, and 1.0 along the objective space.

\subsection{Humanoid Control Task}
We focus on two humanoid control tasks in this paper: \textit{Fall Recovery} and \textit{Walking}. Both tasks rely solely on proprioceptive information.
At timestep $t$, the state is defined as
\begin{equation}
    s_t = [\omega_t, g_t, q_t, \dot{q}_t, a_{t-1}, \eta],
\end{equation}
where $\omega_t \in \mathbb{R}^3$ is the base angular velocity, $g_t \in \mathbb{R}^3$ denotes the gravity vector in the base frame, $q_t$ and $\dot{q}_t$ are the joint positions and velocities, $a_{t-1}$ is the previous action, $\eta \in (0,1]$ is a action scaling factor. For policy learning, both the actor and critic receive as input a combination of five history states and current state, which provides temporal context for decision making in both tasks. The predicted action $a_t$ is mapped to target joint positions through $q_t^{\text{target}} = q^{\text{default}} + \eta a_t$, where $q^{\text{default}}$ is the nominal joint configuration. The corresponding motor torques are then computed by a joint-level PD controller as
\begin{equation}
    \tau_t = K_p \cdot (q_t^{\text{target}} - q_t) - K_d \cdot \dot{q}_t,
\end{equation}
with $K_p$ and $K_d$ denoting the stiffness and damping gains, respectively.

For the reward function design of the two tasks, we referred to prior work \cite{huang2025host, wang2025more, ren2025vb} and additionally incorporated reward terms related to multiple objectives, which are described in \S\ref{sssec:humanoid_walking} and \S\ref{sssec:humanoid_standing}.
To encourage natural behaviors, we also adopt the Adversarial Motion Prior (AMP)~\cite{escontrela2022adversarial}-based reward during the training process. AMP provides a style reward $r^{\text{style}}$ by training a discriminator $D_\phi$ to distinguish between reference motion data and policy-generated trajectories. The discriminator operates on an AMP state sequence $\tau_t = [s^{\text{amp}}_{t-4}, s^{\text{amp}}_{t-3}, \dots, s^{\text{amp}}_t]$, where each AMP state $s^{\text{amp}}_t \in \mathbb{R}^{20}$ is constructed from the 20 joint positions. The discriminator is trained with the following objective:
\begin{align}
\arg\max_{\phi} \ \ & \mathbb{E}_{\tau \sim \mathcal{M}}[(D_{\phi}(\tau) - 1)^2] + \mathbb{E}_{\tau \sim \mathcal{P}}[(D_{\phi}(\tau) + 1)^2] \nonumber \\
& + \frac{\alpha^{d}}{2} \mathbb{E}_{\tau \sim \mathcal{M}}[\|\nabla_{\phi} D_{\phi}(\tau)\|_2],
\end{align}
where $\mathcal{M}$ is the reference dataset, $\mathcal{P}$ contains policy rollouts, and $\alpha^d$ is a regularization coefficient. The discriminator output $D_\phi(\tau_t) \in \mathbb{R}$ is then converted into a smooth reward:
\begin{equation}
    r^{\text{style}}(\tau_t) = c \cdot \max \Big(0,\, 1 - \tfrac{1}{4}(D_\phi(\tau_t) - 1)^2 \Big),
\end{equation}
where $c$ is a manually specified scaling factor. 
In our setup, each task employs a separate discriminator trained on its corresponding reference motion dataset: for \textit{Walking}, we adopt the motions from the LAFAN1 Retargeting Dataset, while for \textit{Fall Recovery} task, we use motions captured and retargeted by ourselves.

\begin{table*}[t]
\centering
\caption{Performance evaluation on the fall recovery task.}
\resizebox{\textwidth}{!}{
\begin{tabular}{c|ccc|ccc|cc}
\toprule
\textbf{Objective} & 
\multicolumn{6}{c|}{\textbf{Energy Efficiency}} & 
\multicolumn{2}{c}{\textbf{Disturbance Robustness}} \\
\cmidrule(l){1-1} \cmidrule(l){2-4} \cmidrule(l){5-7} \cmidrule(l){8-9}
\diagbox{Preference}{Metrics} & \makecell{Avg. Energy \\ Overall (J)} & \makecell{Avg. Energy \\ Fall Recovery (J)} & \makecell{Avg. Energy \\ Recovery (J)}
 & \makecell{Avg. Torque \\ Overall (Nm)} & \makecell{Avg. Torque \\ Fall Recovery (Nm)} & \makecell{Avg. Torque \\ Recovery (Nm)} 
 & \makecell{COM \\ displacement (m)} & \makecell{Success \\ rate (\%)}  \\
\midrule
\multicolumn{9}{c}{\textbf{PCHC (ours)}} \\
\midrule
(1.00, 0.00) & \textbf{463.45} & \textbf{743.14} & \textbf{416.96} & \textbf{223.90} & \textbf{223.01} & \textbf{224.17} & 1.82 & 62.24 \\
(0.75, 0.25) & 480.58 & 747.43 & 435.15 & 241.31 & 227.86 & 245.31 & 1.73 & 67.95 \\
(0.50, 0.50) & 517.90 & 787.09 & 476.94 & 243.04 & 227.26 & 246.00 & 1.62 & 75.73 \\
(0.25, 0.75) & 583.31 & 807.77 & 550.13 & 249.08 & 229.82 & 251.93 & 1.12 & 91.95 \\
(0.00, 1.00) & 666.24 & 790.43 & 646.83 & 256.23 & 232.33 & 260.55 & \textbf{0.75} & \textbf{96.56} \\
\midrule
\multicolumn{9}{c}{\textbf{HoST}} \\
\midrule
 & 550.15 & 745.25 & 467.67 & 235.65 & 224.02 & 240.18 & 1.69 & 69.22 \\
\bottomrule
\end{tabular}}
\label{tab:standup_results}
\end{table*}

\begin{table}[t]
\centering
\caption{Performance evaluation on the locomotion task.} 
\resizebox{\columnwidth}{!}{
\begin{tabular}{c|cc}
\toprule
\diagbox{Preference}{Metrics} & Avg. Stride & Traj. Deviation \\
\midrule
(0.00, 1.00) & \textbf{0.79} & 0.85 \\
(0.25, 0.75) & 0.72 & 0.51 \\
(0.50, 0.50) & 0.67 & 0.33 \\
(0.75, 0.25) & 0.55 & 0.29 \\
(0.10, 0.00) & 0.42 & \textbf{0.23} \\
\midrule
Baseline & 0.59 & 0.46 \\
\bottomrule
\end{tabular}}
\label{tab:loco_results}
\end{table}

\subsubsection{Humanoid Walking}
\label{sssec:humanoid_walking}
In this task, the policy controls 20 actuated joints of the humanoid. Besides the proprioceptive information and preference vector, the velocity commands $\mathbf{v}_{\text{cmd}} \in \mathbb{R}^3$ are also provided as part of the observation input. We define two conflicting objectives for this task: \textbf{stride length} and \textbf{disturbance resistance}, which are quantified by their corresponding reward functions as follows:
\begin{equation}
r_{\text{stride}} = \mathbb{I}_{\text{contact}} \|\mathbf{p}_{\text{current}} - \mathbf{p}_{\text{last\_contact}}\|_2
\end{equation}
\begin{equation}
r_{\text{disturb}} =  \mathbb{I}_{\text{after-disturb}} (- \|\mathbf{x}_{\text{base}} - \mathbf{x}_{\text{last}}\|_2 - \big| \mathbf{v}_{\text{base}} \cdot \hat{\mathbf{d}} \, \big|) 
\end{equation}
where $\mathbb{I}_{\text{contact}}$ and $\mathbb{I}_{\text{after-disturb}}$ are indicator functions used to determine whether a foot is in contact and whether the current step is within a period of time after a disturbance, $\mathbf{p}_{\text{current}}$ and $\mathbf{p}_{\text{last\_contact}}$ is the foot pos, $\mathbf{x}_{\text{base}}$ and $\mathbf{x}_{\text{last}}$ denote the horizontal base positions before and after an external disturbance,  $\mathbf{v}_{\text{base}}$ is linear velocity of the base, $\hat{\mathbf{d}}$ is unit vector in the disturbance direction. The stride length objective encourages the policy to take larger steps. In contrast, disturbance resistance emphasizes stability and robustness against disturbance, often resulting in shorter and more conservative steps. Through preference-conditioned optimization, our framework enables the humanoid to modulate its gait to favor either longer strides or stronger robustness.

\subsubsection{Humanoid Fall Recovery}
\label{sssec:humanoid_standing}
The \textit{Fall Recovery} task trains the humanoid to rise from a prone posture. Here, the policy controls 23 actuated joints. Similar to the walking task, we consider \textbf{disturbance resistance} and \textbf{energy efficiency} as the multi-objective setting. The energy efficiency reward function is designed as follows:
\begin{equation}
r_{\text{energy}} = -\kappa_\tau \sum_{j=1}^{N_\text{dof}} \tau_j^2
                     - \kappa_p \sum_{j=1}^{N_\text{dof}} |\dot{q}_j| \, |\tau_j|,
\end{equation}
where $N_\text{dof}$ is the number of actuated joints, $\dot{q}_j$ and $\tau_j$ are the velocity and torque of joint $j$, and $\kappa_\tau, \kappa_p > 0$ are scaling coefficients. The disturbance resistance reward function is same as the walking task. Disturbance resistance requires the policy to maintain robustness against external disturbance applied during both the recovery phase and the subsequent standing posture. Energy efficiency, on the other hand, encourages the policy to minimize torque consumption through the whole process. These two objectives are inherently conflicting since stronger disturbance resistance typically demands higher torque expenditure. Note that excessive coefficients in $r_{\text{energy}}$ may hinder the robot from standing.

\section{Result and Discussion}
We conduct parallel training using the Isaac Gym simulator. 
Each training iteration aggregates experience from 4096 parallel environments, with each episode lasting 10 seconds. The entire training process was performed on a single NVIDIA GeForce RTX 4090 GPU. For both the walking and fall recovery tasks, each policy was trained for a total of 12000 iterations. The control policy operates at 50 Hz and outputs target joint positions, which are tracked by a PD controller running at 500 Hz to compute the corresponding joint torques. The policy is trained and deployed specifically for the Unitree G1 humanoid robot. To ensure efficient real-time inference, the model is converted to the Open Neural Network Exchange (ONNX) format and executed on the robot's onboard computer.

\subsection{Simulation Result}

\begin{figure*}[t]
  \centering
  \begin{subfigure}{0.50\linewidth}
    \centering
    \includegraphics[width=\linewidth]{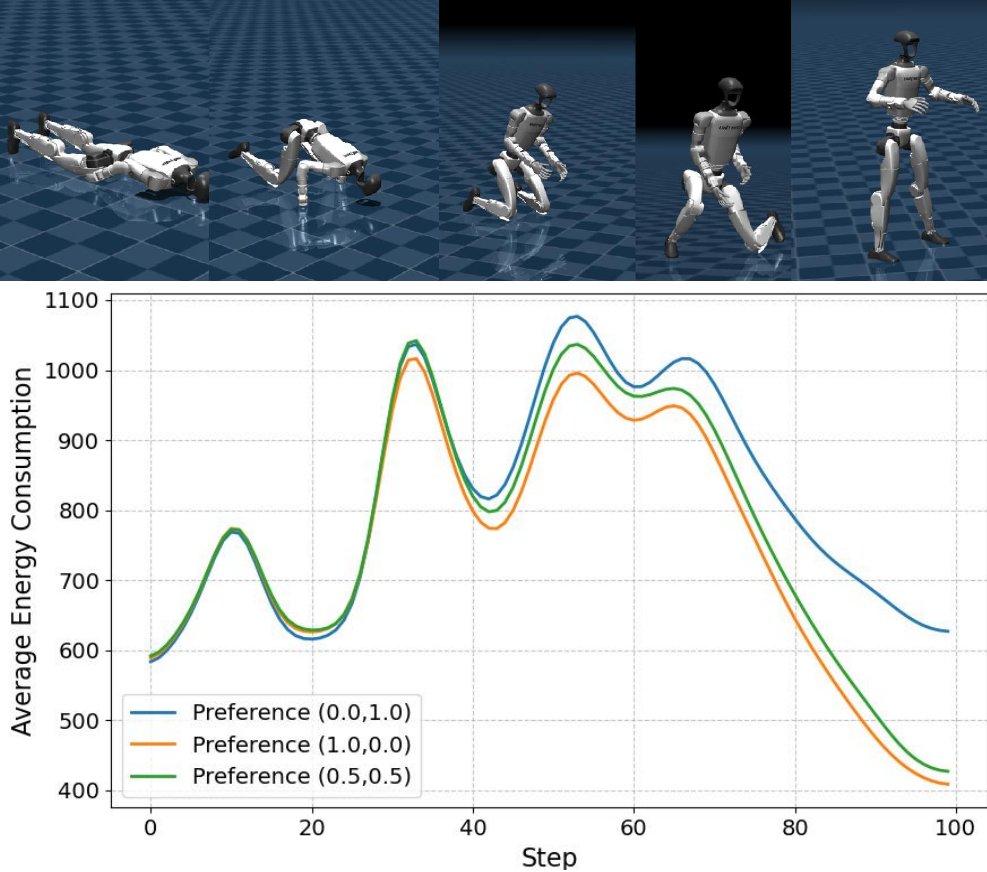}
    \caption{Fall Recovery task results.} 
    \label{fig:standupb}
  \end{subfigure}
  \hfill
  \begin{subfigure}{0.46\linewidth}
    \centering
    \includegraphics[width=\linewidth]{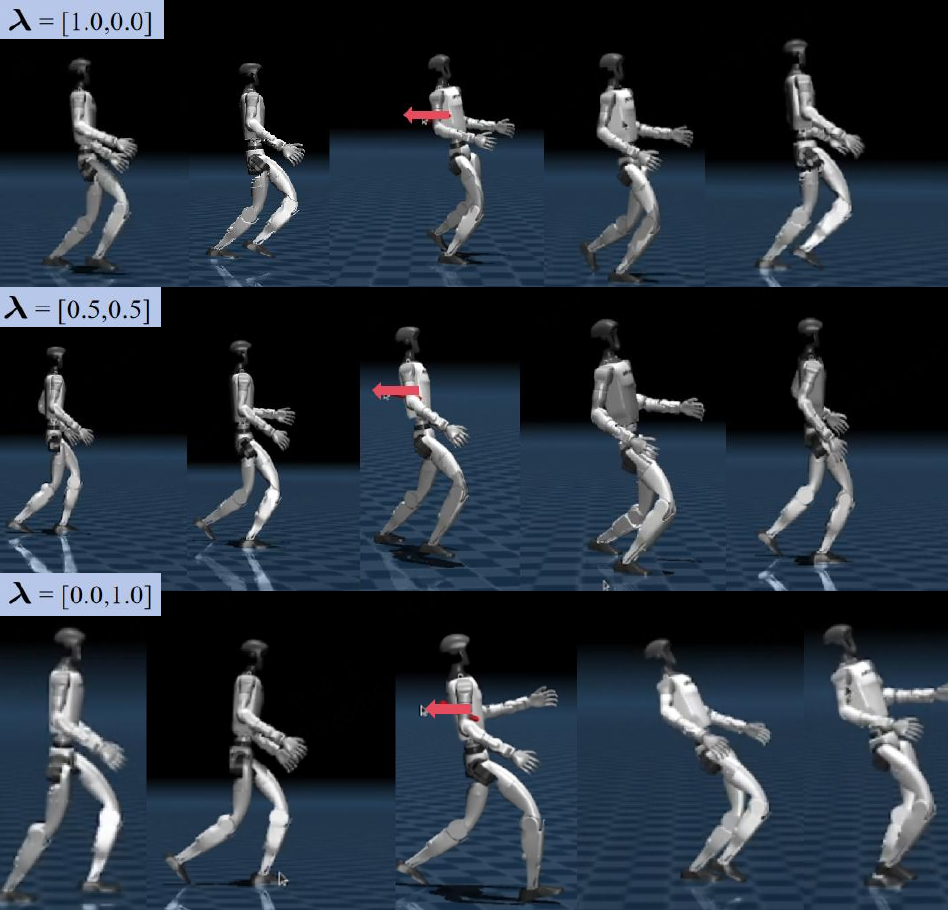}
    \caption{Walking task results.} 
    \label{fig:walking_sub}
  \end{subfigure}

  \caption{The performance of our policy with different preference vector $\bm \lambda$ on \textit{Fall Recovery} and \textit{Walking} task.}
  \vspace{-1.5em}
  \label{fig:simulation}
\end{figure*}

\begin{figure}[t]
  \centering
  \includegraphics[width=\linewidth]{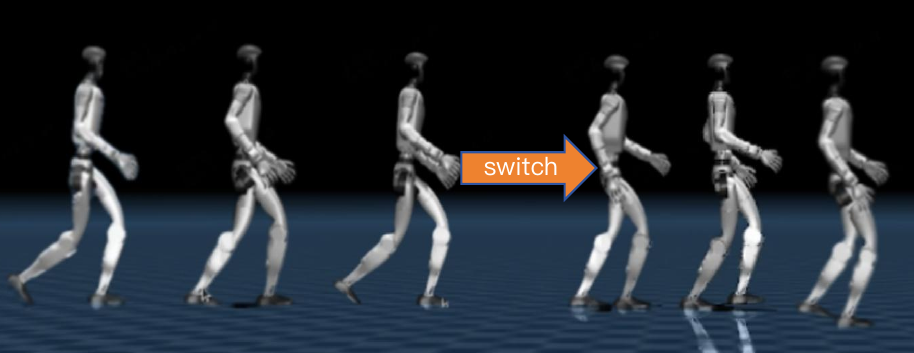}
  \caption{Demonstration of dynamic preference switching during task execution. The preference vector $\bm \lambda$ is instantaneously adjusted from [0.0, 1.0] to [1.0, 0.0].}
  \vspace{-1.5em}
  \label{fig:dynamic}
\end{figure}

\paragraph{Fall Recovery task}
We first evaluate \textit{PCHC} on the humanoid fall recovery task, where the robot starts from a prone posture. To study how preferences affect the learned behaviors, we vary the preference vector $\bm \lambda = (\lambda_1, \lambda_2)$ across five settings, ranging from $\lambda=(1.0,0.0)$, which prioritizes energy efficiency, to $\bm \lambda=(0.0,1.0)$, which emphasizes robustness against disturbance. Each policy is evaluated in 1000 parallel environments. We report results across multiple metrics: (1) \textbf{Average energy consumption}, reported in Joules (J), including both the overall value across the entire process and the decomposed values for the recovery phase and the standing phase; (2) \textbf{Average joint torque}, measured in Newton–meters (Nm), also reported as an overall value and further separated into the recovery and standing phases; (3) Disturbance robustness under small disturbance, measured by the \textbf{COM displacement} in meters(m) after mild pushes; and (4) \textbf{Success rate} defined as the percentage (\%) of trials in which the robot remains standing without falling when subjected to strong external disturbances. Baseline is HoST \cite{standup} augmented with AMP motion priors.

As shown in Table~\ref{tab:standup_results}, \textit{PCHC} effectively adapts its behavior according to the preference vector. When prioritizing energy efficiency, the policy exhibits reduced robustness and success rate under perturbations. Conversely, when robustness is emphasized, the policy maintains stable standing under large disturbances with increased energy consumption. Intermediate preferences yield smooth trade-offs between these two extremes, producing balanced solutions. In addition, we record the \textbf{step-wise energy consumption} throughout the entire fall recovery process (Fig.~\ref{fig:simulation}(a)), comparing three representative preferences. The curves show how different preference settings modulate energy usage, further validating the role of preference conditioning in shaping task behaviors.

From the perspective of Pareto front (PF) analysis, HoST corresponds to a single fixed solution, which is strictly dominated by multiple solutions obtained through \textit{PCHC}. In other words, for the same or lower energy consumption, \textit{PCHC} achieves higher robustness, while for the same robustness, it yields lower energy consumption. This shows that the baseline solution lies inside the dominated region, whereas \textit{PCHC} spans a broader set of non-dominated points approximating the Pareto front.These results show that preference conditioning enables \textit{PCHC} to flexibly navigate the trade-off between energy consumption and disturbance resistance in the fall recovery task. By adjusting the preference vector, the policy smoothly interpolates between behaviors, showing that the baseline lies within the dominated region.

\paragraph{Walking task.}
We further evaluate the locomotion performance of \textit{PCHC} on the walking task, where the humanoid is commanded to walk forward on flat terrain. Similar to the fall recovery task, to examine how preference conditioning influences gait behavior, we vary the preference vector $\bm \lambda = [\lambda_1, \lambda_2]$ across five setting for comparison. Each trained policy is tested in 1000 parallel rollouts, and we record the average \textbf{stride length} and \textbf{trajectory deviation} both expressed in meters (m). For reference, we include a baseline policy that follows the design of prior approaches \cite{huang2025host, wang2025more, gu2024advancing}.

As summarized in Table~\ref{tab:loco_results}, different preferences lead to different locomotion styles. When stride length is prioritized ($\bm \lambda=[0.0,1.0]$), the robot takes larger steps but suffers from increased trajectory deviation. Conversely, when stability is emphasized ($\bm\lambda=[1.0,0.0]$), the policy achieves minimal trajectory deviation but at the cost of shorter strides. Besides, the results showing that intermediate preferences produce gradual transitions, indicating that the policy flexibly balances the two objectives. Similarly, Fig.~\ref{fig:simulation}(b) illustrates representative simulation results under three preference settings, providing a direct visualization of how preferences modulate the trade-off between stride length and stability.

Compared to the baseline, which remains confined to a single operating point, \textit{PCHC} covers a much broader portion of the stride–stability spectrum. Instead of being restricted to a fixed compromise, the preference-conditioned policies provide a continuous set of solutions: some achieve longer strides with larger deviations, while others ensure highly stable trajectories at the cost of reduced stride length. This diversity demonstrates that preference conditioning enables the locomotion policy to adapt flexibly to different requirements, whereas the baseline remains dominated by multiple solutions obtained through \textit{PCHC}.

To evaluate the robustness of our framework under changing requirements, we further tested the policy's ability to handle dynamic preference switching during task execution. As shown in Fig.~\ref{fig:dynamic}, the preference vector $\boldsymbol{\lambda}$ is instantaneously altered while the humanoid is executing the task. The policy demonstrates the capability to seamlessly transition its behavior in real-time without any instability or failure. This confirms that the PCHC framework is not limited to static preference settings, but can dynamically and safely adapt to evolving objective priorities in complex environments.

\begin{figure}[t]
  \centering
  \includegraphics[width=\linewidth]{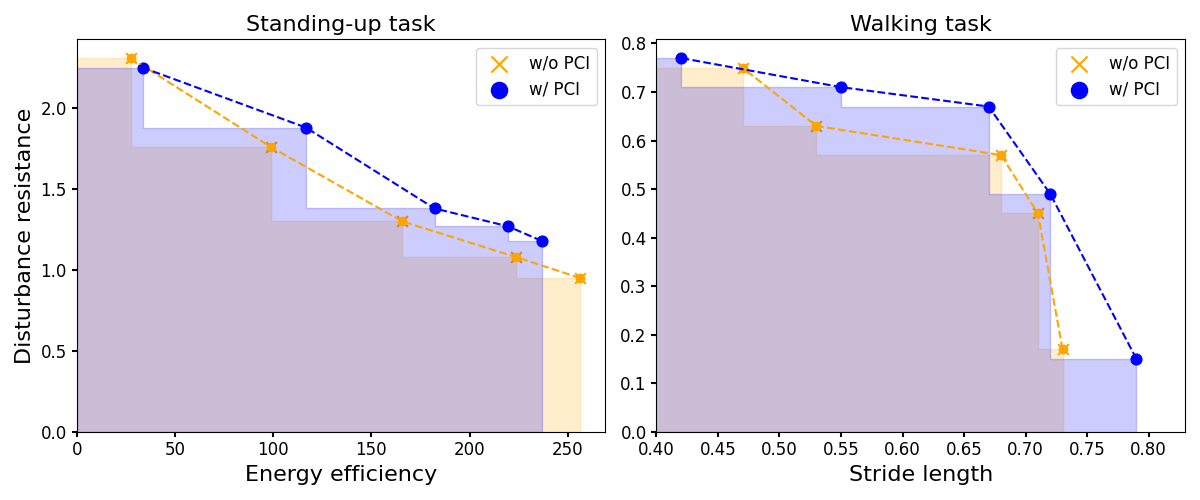}
  \caption{Ablation study comparing our PCI module (w/ PCI) against a standard trainable gating network (w/o PCI).}
  \label{fig:ablation_pf}
\end{figure}

\begin{figure}[t]
\centering
\includegraphics[width=1.0\linewidth, height=0.3\textheight]{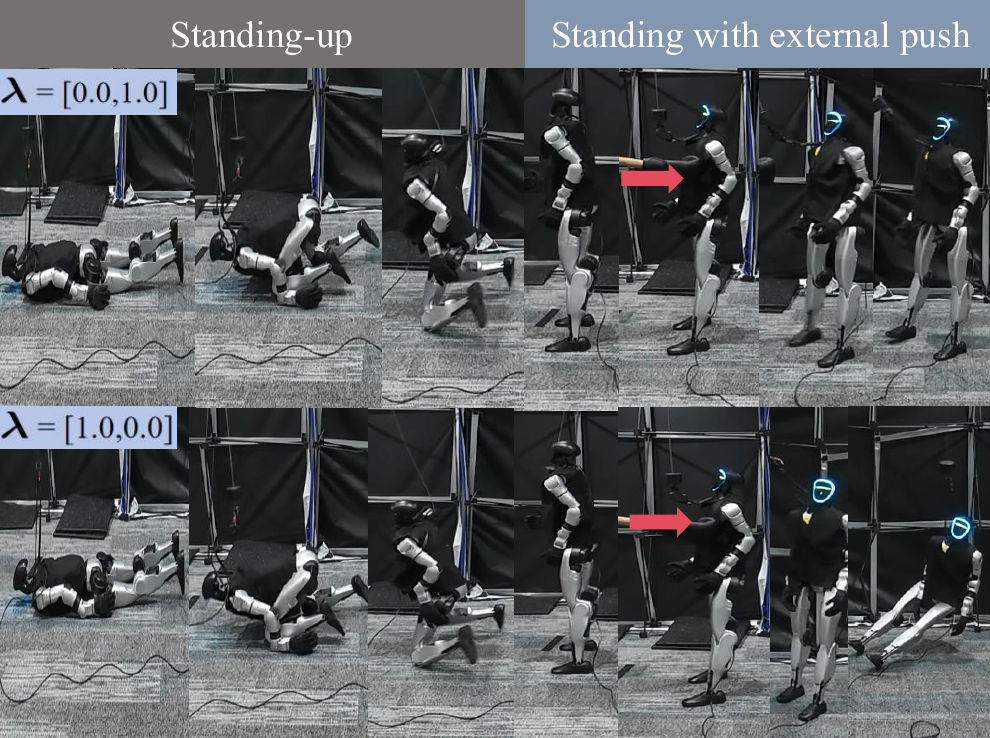}
\caption{Real world experiments results of fall recovery task on a Unitree G1 robot.}
\label{fig:host}
\vspace{-1.5em}
\end{figure}

\paragraph{Ablation on Preference Condition Injection}
To evaluate the effectiveness of the proposed Preference Condition Injection (PCI) module, we compare it against a standard trainable gating network. As illustrated by the Pareto front (PF) curves in Fig.~\ref{fig:ablation_pf}, removing the PCI module leads to a noticeable contraction of the front both in the fall recovery task and walking task. To quantify these differences, we compute the \textbf{hypervolume} and \textbf{sparsity} of the obtained PFs. On the fall recovery task, the PCI-enabled policy achieves a hypervolume of 313.78 compared to 306.02 without PCI, with sparsity reduced from 3493.85 to 3210.61. On the walking task, the hypervolume increases from 0.14 to 0.20, while sparsity shows little change. These results confirm that our Beta-distribution based routing provides superior coverage of the objective space and a more faithful approximation of the Pareto front than a conventional gating architecture.

\begin{figure}[t]
  \centering
  \includegraphics[width=\linewidth]{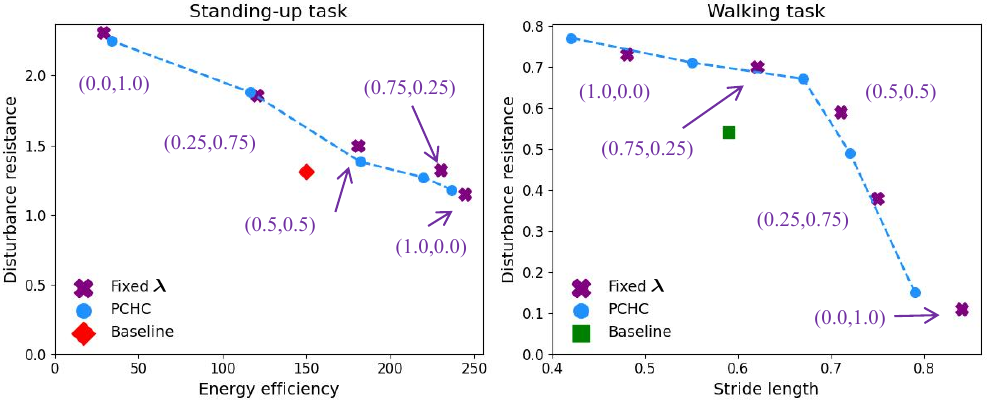}
  \caption{Objective space evaluation of PCHC, Fixed $\boldsymbol{\lambda}$ policies, and Baseline. }
  \label{fig:fixed_pref}
\end{figure}

\paragraph{Compared with fixed preference}
During training, \textit{PCHC} randomly samples a preference vector $\bm \lambda$ and concatenates it with the robot’s observation, so that a single policy can adapt to different trade-offs. An important concern is whether introducing preference conditioning would compromise the performance at individual trade-off points compared to policies trained with fixed preference vector. To verify this, we trained fixed-preference policies for both tasks, each with a specific preference vector $\bm \lambda$. As shown in Fig.~\ref{fig:fixed_pref}, the results demonstrate that \textit{PCHC} achieves comparable performance at the corresponding preferences. This indicates that Preference conditioning retains fixed-preference training effectiveness and extends it to a continuous solution set in a single policy.

\subsection{Real-world Result}
We deploy the trained \textit{PCHC} policies onto a Unitree G1 humanoid robot to directly examine their sim-to-real transfer under different preferences.

\paragraph{fall recovery task}
This task involves a trade-off between energy efficiency and robustness to external pushes. As illustrated in Fig.~\ref{fig:host}, with $\bm\lambda=[1.0,0.0]$ the robot rises with relatively low torque effort but quickly loses balance when disturbed. In contrast, with $\bm \lambda=[0.0,1.0]$ the robot generates stronger and more stable motions that enable it to withstand the same perturbations, though at the cost of higher joint effort.

\paragraph{Walking task}
For locomotion, the objectives are stride length and stability. As shown in Fig.~\ref{fig:walking_performance}, under $\bm\lambda=[0.0,1.0]$ the robot achieves longer strides but shows larger trajectory deviation when external forces are applied. Conversely, with $\bm\lambda=[1.0,0.0]$ the robot maintains better disturbance resistance, albeit with shorter steps.

\section{Conclusion}
In this paper, we propose \textit{PCHC} framework which enables a single policy to perform preference-aligned humanoid behaviors. By leveraging the Beta distribution parameterized by the preference vector, the \textit{PCI} module efficiently enables the policy to satisfy specified objective trade-offs. However, our framework is primarily applicable to tasks with two objectives, whereas many conflicting objectives in humanoid tasks require scenario-dependent consideration. Moreover, our approach has been applied only to blind humanoid control tasks, without integrating external sensors. Future work will consider a broader range of objectives requiring trade-offs and vision-based tasks.

\bibliographystyle{IEEEtran}
\bibliography{main}

@article{huang2025host,
  title={Learning humanoid standing-up control across diverse postures},
  author={Huang, Tao and Ren, Junli and Wang, Huayi and Wang, Zirui and Ben, Qingwei and Wen, Muning and Chen, Xiao and Li, Jianan and Pang, Jiangmiao},
  journal={arXiv preprint arXiv:2502.08378},
  year={2025}
}

@inproceedings{standup,
  title={Learning Getting-Up Policies for Real-World Humanoid Robots},
  author={He, Xialin and Dong, Runpei and Chen, Zixuan and Gupta, Saurabh},
  booktitle={Robotics: Science and Systems},
  year={2025}
}

@article{zhuang2024humanoid,
  title={Humanoid parkour learning},
  author={Zhuang, Ziwen and Yao, Shenzhe and Zhao, Hang},
  journal={arXiv preprint arXiv:2406.10759},
  year={2024}
}

@article{fu2024humanplus,
  title={Humanplus: Humanoid shadowing and imitation from humans},
  author={Fu, Zipeng and Zhao, Qingqing and Wu, Qi and Wetzstein, Gordon and Finn, Chelsea},
  journal={arXiv preprint arXiv:2406.10454},
  year={2024}
}

@article{gu2024humanoid,
  title={Humanoid-gym: Reinforcement learning for humanoid robot with zero-shot sim2real transfer},
  author={Gu, Xinyang and Wang, Yen-Jen and Chen, Jianyu},
  journal={arXiv preprint arXiv:2404.05695},
  year={2024}
}

@article{ren2025vb,
  title={Vb-com: Learning vision-blind composite humanoid locomotion against deficient perception},
  author={Ren, Junli and Huang, Tao and Wang, Huayi and Wang, Zirui and Ben, Qingwei and Long, Junfeng and Yang, Yanchao and Pang, Jiangmiao and Luo, Ping},
  journal={arXiv preprint arXiv:2502.14814},
  year={2025}
}

@inproceedings{long2025learning,
  title={Learning humanoid locomotion with perceptive internal model},
  author={Long, Junfeng and Ren, Junli and Shi, Moji and Wang, Zirui and Huang, Tao and Luo, Ping and Pang, Jiangmiao},
  booktitle={2025 IEEE International Conference on Robotics and Automation (ICRA)},
  pages={9997--10003},
  year={2025},
  organization={IEEE}
}

@article{wang2025beamdojo,
  title={Beamdojo: Learning agile humanoid locomotion on sparse footholds},
  author={Wang, Huayi and Wang, Zirui and Ren, Junli and Ben, Qingwei and Huang, Tao and Zhang, Weinan and Pang, Jiangmiao},
  journal={arXiv preprint arXiv:2502.10363},
  year={2025}
}

@article{gu2024advancing,
  title={Advancing humanoid locomotion: Mastering challenging terrains with denoising world model learning},
  author={Gu, Xinyang and Wang, Yen-Jen and Zhu, Xiang and Shi, Chengming and Guo, Yanjiang and Liu, Yichen and Chen, Jianyu},
  journal={arXiv preprint arXiv:2408.14472},
  year={2024}
}

@inproceedings{cui2024adapting,
  title={Adapting humanoid locomotion over challenging terrain via two-phase training},
  author={Cui, Wenhao and Li, Shengtao and Huang, Huaxing and Qin, Bangyu and Zhang, Tianchu and Zheng, Liang and Tang, Ziyang and Hu, Chenxu and Yan, NING and Chen, Jiahao and others},
  booktitle={8th Annual Conference on Robot Learning},
  year={2024}
}

@article{PPO1,
  title={Proximal policy optimization algorithms},
  author={Schulman, John and Wolski, Filip and Dhariwal, Prafulla and Radford, Alec and Klimov, Oleg},
  journal={arXiv preprint arXiv:1707.06347},
  year={2017}
}

@inproceedings{PPO2,
  title={Implementation matters in deep rl: A case study on ppo and trpo},
  author={Engstrom, Logan and Ilyas, Andrew and Santurkar, Shibani and Tsipras, Dimitris and Janoos, Firdaus and Rudolph, Larry and Madry, Aleksander},
  booktitle={ICLR},
  year={2019}
}

@article{kaelbling1996reinforcement,
  title={Reinforcement learning: A survey},
  author={Kaelbling, Leslie Pack and Littman, Michael L and Moore, Andrew W},
  journal={Journal of artificial intelligence research},
  volume={4},
  pages={237--285},
  year={1996}
}

@misc{gao2024mpc,
      title={Time-Varying Foot-Placement Control for Underactuated Humanoid Walking on Swaying Rigid Surfaces}, 
      author={Yuan Gao and Victor Paredes and Yukai Gong and Zijian He and Ayonga Hereid and Yan Gu},
      year={2024},
      eprint={2409.08371},
      archivePrefix={arXiv},
      primaryClass={cs.RO},
      url={https://arxiv.org/abs/2409.08371}, 
}

@article{gu2025humanoid,
  title={Humanoid locomotion and manipulation: Current progress and challenges in control, planning, and learning},
  author={Gu, Zhaoyuan and Li, Junheng and Shen, Wenlan and Yu, Wenhao and Xie, Zhaoming and McCrory, Stephen and Cheng, Xianyi and Shamsah, Abdulaziz and Griffin, Robert and Liu, C Karen and others},
  journal={arXiv preprint arXiv:2501.02116},
  year={2025}
}

@article{kim2007humanoidmpc,
  title={Walking control algorithm of biped humanoid robot on uneven and inclined floor},
  author={Kim, Jung-Yup and Park, Ill-Woo and Oh, Jun-Ho},
  journal={Journal of intelligent and robotic systems},
  volume={48},
  number={4},
  pages={457--484},
  year={2007},
  publisher={Springer}
}

@article{wang2025more,
  title={MoRE: Mixture of Residual Experts for Humanoid Lifelike Gaits Learning on Complex Terrains},
  author={Wang, Dewei and Wang, Xinmiao and Liu, Xinzhe and Shi, Jiyuan and Zhao, Yingnan and Bai, Chenjia and Li, Xuelong},
  journal={arXiv preprint arXiv:2506.08840},
  year={2025}
}

@article{shi2025adversarial,
  title={Adversarial Locomotion and Motion Imitation for Humanoid Policy Learning},
  author={Shi, Jiyuan and Liu, Xinzhe and Wang, Dewei and Lu, Ouyang and Schwertfeger, S{\"o}ren and Sun, Fuchun and Bai, Chenjia and Li, Xuelong},
  journal={arXiv preprint arXiv:2504.14305},
  year={2025}
}

@article{ben2025homie,
  title={Homie: Humanoid loco-manipulation with isomorphic exoskeleton cockpit},
  author={Ben, Qingwei and Jia, Feiyu and Zeng, Jia and Dong, Junting and Lin, Dahua and Pang, Jiangmiao},
  journal={arXiv preprint arXiv:2502.13013},
  year={2025}
}

@article{xue2025unified,
  title={A Unified and General Humanoid Whole-Body Controller for Versatile Locomotion},
  author={Xue, Yufei and Dong, Wentao and Liuˆ, Minghuan and Zhang, Weinan and Pang, Jiangmiao},
  journal={arXiv preprint arXiv:2502.03206},
  year={2025}
}

@article{zakka2025mujoco,
  title={Mujoco playground},
  author={Zakka, Kevin and Tabanpour, Baruch and Liao, Qiayuan and Haiderbhai, Mustafa and Holt, Samuel and Luo, Jing Yuan and Allshire, Arthur and Frey, Erik and Sreenath, Koushil and Kahrs, Lueder A and others},
  journal={arXiv preprint arXiv:2502.08844},
  year={2025}
}

@article{makoviychuk2021isaac,
  title={Isaac gym: High performance gpu-based physics simulation for robot learning},
  author={Makoviychuk, Viktor and Wawrzyniak, Lukasz and Guo, Yunrong and Lu, Michelle and Storey, Kier and Macklin, Miles and Hoeller, David and Rudin, Nikita and Allshire, Arthur and Handa, Ankur and others},
  journal={arXiv preprint arXiv:2108.10470},
  year={2021}
}

@article{hayes2021practical,
  title={A practical guide to multi-objective reinforcement learning and planning},
  author={Hayes, Conor F and R{\u{a}}dulescu, Roxana and Bargiacchi, Eugenio and K{\"a}llstr{\"o}m, Johan and Macfarlane, Matthew and Reymond, Mathieu and Verstraeten, Timothy and Zintgraf, Luisa M and Dazeley, Richard and Heintz, Fredrik and others},
  journal={arXiv preprint arXiv:2103.09568},
  year={2021}
}

@inproceedings{xu2020prediction,
  title={Prediction-guided multi-objective reinforcement learning for continuous robot control},
  author={Xu, Jie and Tian, Yunsheng and Ma, Pingchuan and Rus, Daniela and Sueda, Shinjiro and Matusik, Wojciech},
  booktitle={International conference on machine learning},
  pages={10607--10616},
  year={2020},
  organization={PMLR}
}

@article{cai2023distributional,
  title={Distributional pareto-optimal multi-objective reinforcement learning},
  author={Cai, Xin-Qiang and Zhang, Pushi and Zhao, Li and Bian, Jiang and Sugiyama, Masashi and Llorens, Ashley},
  journal={Advances in Neural Information Processing Systems},
  volume={36},
  pages={15593--15613},
  year={2023}
}

@article{terekhov2024search,
  title={In Search for Architectures and Loss Functions in Multi-Objective Reinforcement Learning},
  author={Terekhov, Mikhail and Gulcehre, Caglar},
  journal={arXiv preprint arXiv:2407.16807},
  year={2024}
}

@article{reymond2022pareto,
  title={Pareto conditioned networks},
  author={Reymond, Mathieu and Bargiacchi, Eugenio and Now{\'e}, Ann},
  journal={arXiv preprint arXiv:2204.05036},
  year={2022}
}

@article{rame2023rewarded,
  title={Rewarded soups: towards pareto-optimal alignment by interpolating weights fine-tuned on diverse rewards},
  author={Rame, Alexandre and Couairon, Guillaume and Dancette, Corentin and Gaya, Jean-Baptiste and Shukor, Mustafa and Soulier, Laure and Cord, Matthieu},
  journal={Advances in Neural Information Processing Systems},
  volume={36},
  pages={71095--71134},
  year={2023}
}

@article{jang2023personalized,
  title={Personalized soups: Personalized large language model alignment via post-hoc parameter merging},
  author={Jang, Joel and Kim, Seungone and Lin, Bill Yuchen and Wang, Yizhong and Hessel, Jack and Zettlemoyer, Luke and Hajishirzi, Hannaneh and Choi, Yejin and Ammanabrolu, Prithviraj},
  journal={arXiv preprint arXiv:2310.11564},
  year={2023}
}

@article{van2014multi,
  title={Multi-objective reinforcement learning using sets of pareto dominating policies},
  author={Van Moffaert, Kristof and Now{\'e}, Ann},
  journal={The Journal of Machine Learning Research},
  volume={15},
  number={1},
  pages={3483--3512},
  year={2014},
  publisher={JMLR. org}
}

@article{navon2020learning,
  title={Learning the pareto front with hypernetworks},
  author={Navon, Aviv and Shamsian, Aviv and Chechik, Gal and Fetaya, Ethan},
  journal={arXiv preprint arXiv:2010.04104},
  year={2020}
}

@article{basaklar2022pd,
  title={Pd-morl: Preference-driven multi-objective reinforcement learning algorithm},
  author={Basaklar, Toygun and Gumussoy, Suat and Ogras, Umit Y},
  journal={arXiv preprint arXiv:2208.07914},
  year={2022}
}

@article{yang2019generalized,
  title={A generalized algorithm for multi-objective reinforcement learning and policy adaptation},
  author={Yang, Runzhe and Sun, Xingyuan and Narasimhan, Karthik},
  journal={Advances in neural information processing systems},
  volume={32},
  year={2019}
}

@inproceedings{alegre2022mo,
  title={MO-Gym: A library of multi-objective reinforcement learning environments},
  author={Alegre, Lucas N and Felten, Florian and Talbi, El-Ghazali and Danoy, Gr{\'e}goire and Now{\'e}, Ann and Bazzan, Ana LC and da Silva, Bruno C},
  booktitle={Proceedings of the 34th Benelux Conference on Artificial Intelligence BNAIC/Benelearn},
  volume={2022},
  pages={2},
  year={2022}
}

@article{AMP,
  title={Amp: Adversarial motion priors for stylized physics-based character control},
  author={Peng, Xue Bin and Ma, Ze and Abbeel, Pieter and Levine, Sergey and Kanazawa, Angjoo},
  journal={ACM Transactions on Graphics (ToG)},
  volume={40},
  number={4},
  year={2021},
  publisher={ACM New York, NY, USA}
}

@inproceedings{escontrela2022adversarial,
  title={Adversarial motion priors make good substitutes for complex reward functions},
  author={Escontrela, Alejandro and Peng, Xue Bin and Yu, Wenhao and Zhang, Tingnan and Iscen, Atil and Goldberg, Ken and Abbeel, Pieter},
  booktitle={2022 IEEE/RSJ International Conference on Intelligent Robots and Systems (IROS)},
  pages={25--32},
  year={2022},
  organization={IEEE}
}

@article{xie2025kungfubot,
        title={KungfuBot: Physics-Based Humanoid Whole-Body Control for Learning Highly-Dynamic Skills},
        author={Xie, Weiji and Han, Jinrui and Zheng, Jiakun and Li, Huanyu and Liu, Xinzhe and Shi, Jiyuan and Zhang, Weinan and Bai, Chenjia and Li, Xuelong},
        journal={arXiv preprint arXiv:2506.12851},
        year={2025}
}

@article{he2025asap,
          title={ASAP: Aligning Simulation and Real-World Physics for Learning Agile Humanoid Whole-Body Skills},
          author={He, Tairan and Gao, Jiawei and Xiao, Wenli and Zhang, Yuanhang and Wang, Zi and Wang, Jiashun and Luo, Zhengyi and He, Guanqi and Sobanbabu, Nikhil and Pan, Chaoyi and Yi, Zeji and Qu, Guannan and Kitani, Kris and Hodgins, Jessica and Fan, Linxi "Jim" and Zhu, Yuke and Liu, Changliu and Shi, Guanya},
          journal={arXiv preprint arXiv:2502.01143},
          year={2025}
}

@inproceedings{mysore2022multi,
  title={Multi-critic actor learning: Teaching rl policies to act with style},
  author={Mysore, Siddharth and Cheng, George and Zhao, Yunqi and Saenko, Kate and Wu, Meng},
  booktitle={International Conference on Learning Representations},
  year={2022}
}

@inproceedings{van_moffaert_scalarized_2013,
    address = {Singapore, Singapore},
    title = {Scalarized multi-objective reinforcement learning: {Novel} design techniques},
    
    booktitle = {2013 {IEEE} {Symposium} on {Adaptive} {Dynamic} {Programming} and {Reinforcement} {Learning} ({ADPRL})},
    publisher = {IEEE},
    author = {Van Moffaert, Kristof and Drugan, Madalina M. and Nowe, Ann},
    year = {2013},
}

@article{schulman2015high,
  title={High-dimensional continuous control using generalized advantage estimation},
  author={Schulman, John and Moritz, Philipp and Levine, Sergey and Jordan, Michael and Abbeel, Pieter},
  journal={arXiv preprint arXiv:1506.02438},
  year={2015}
}

\end{document}